\documentclass[letterpaper, 10 pt, conference]{ieeeconf}

\IEEEoverridecommandlockouts
\usepackage{amsmath}
\usepackage{amssymb}
\usepackage{graphicx}
\usepackage{booktabs}
\usepackage{xcolor}
\usepackage{url}
\usepackage{xspace}
\usepackage{colortbl}
\usepackage{multirow}
\usepackage{caption}
\usepackage{etoolbox}

\definecolor{novelobject}{RGB}{15, 158, 213}
\definecolor{novelscene}{RGB}{77, 167, 46}
\definecolor{novelview}{RGB}{233, 113, 50}

\newcommand{\methodname}{RoboDream\xspace}

\usepackage{gradient-text}

\title{\LARGE \bf
\methodname: \\Compositional World Models for Scalable Robot Data Synthesis
}

\author{
    Junjie Ye\textsuperscript{$1,2$} \quad
    Rong Xue\textsuperscript{$1$} \quad
    Basile Van Hoorick\textsuperscript{$2$} \quad
    Runhao Li\textsuperscript{$1$} \quad
    Harshitha Rajaprakash \textsuperscript{$1$} \quad \\
    Pavel Tokmakov\textsuperscript{$2$} \quad
    Muhammad Zubair Irshad\textsuperscript{$2$} \quad
    Vitor Guizilini\textsuperscript{$2,\dagger$} \quad
    Yue Wang\textsuperscript{$1,\dagger$} \quad \\
    \textsuperscript{$1$}USC Physical Superintelligence (PSI) Lab \quad
    \textsuperscript{$2$}Toyota Research Institute \\
    \textsuperscript{$\dagger$}Equal advising
}

\begin{document}

\newcommand{\insertteaser}{
    \includegraphics[width=0.99\linewidth]{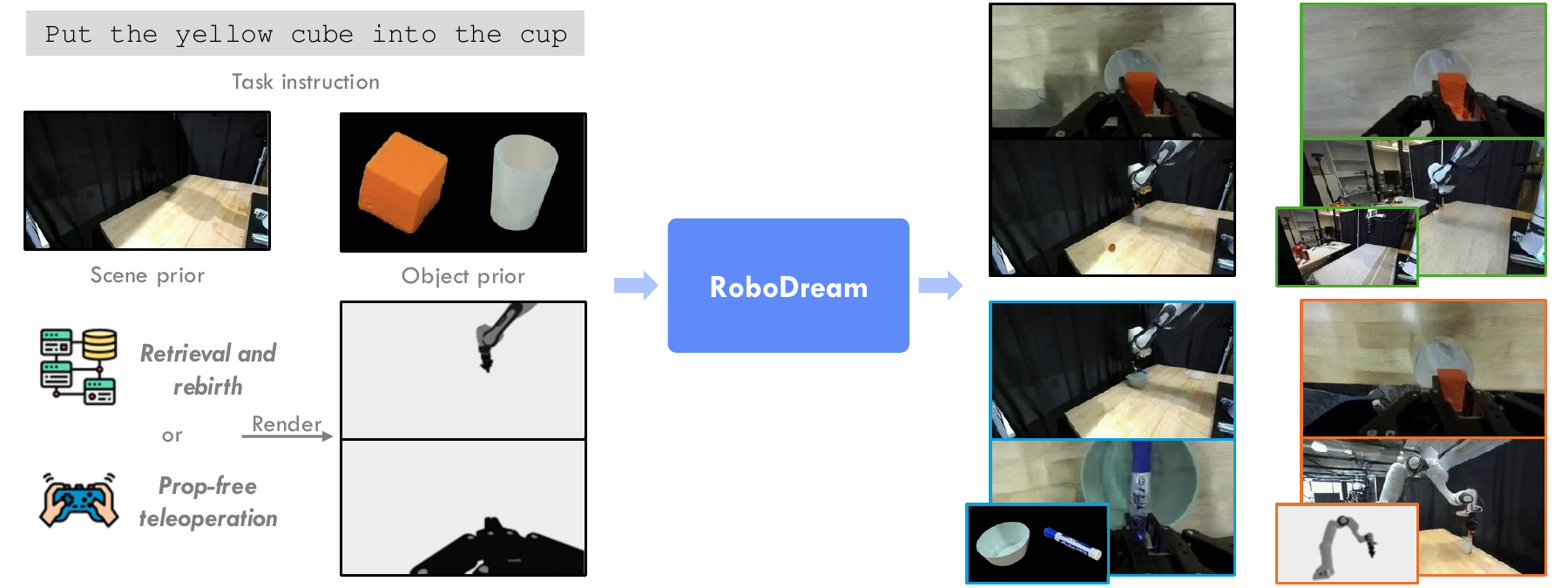}
    \captionof{figure}{%
        \textbf{\methodname} generates robot demonstrations with \textbf{\textcolor{novelobject}{novel objects}}, in \textbf{\textcolor{novelscene}{novel scenes}}, from \textbf{\textcolor{novelview}{novel views}} via \textbf{compositional synthesis}.
        Built on the insight that actions, objects, and scenes are distinct, recombinable elements, we condition video diffusion on three decoupled inputs: a rendered robot-only trajectory, a scene prior (background), and an object prior.
        This framework enables two scalable deployment modes: \emph{retrieval and rebirth},
        which rebirths retrieved trajectories in novel contexts; and \emph{prop-free teleoperation} (in either real world or simulator), where the
        operator manipulates empty air and the model paints in the objects
        afterwards.
        Crucially, these modes achieve such generalization by simply changing the scene or object priors, or by rendering robot motion from a different viewpoint and capturing a corresponding real-world scene prior.
    }
    \label{fig:teaser}
 \vspace{-5pt}
}

\makeatletter
\apptocmd{\@maketitle}{\centering\insertteaser}{}{}
\makeatother

\maketitle
\setcounter{figure}{1}

\thispagestyle{empty}
\pagestyle{empty}

\begin{abstract}

Scaling robot learning requires large-scale, diverse demonstrations, yet
real-world data collection via teleoperation remains prohibitively expensive
and time-consuming.
While video diffusion models offer a promising avenue for data scaling,
existing generative approaches are often limited to superficial visual
augmentation, or suffer from embodiment hallucinations that yield
physically infeasible motions.
We present a generalizable embodiment-centric world model that achieves
scalable data generation by synthesizing photorealistic demonstrations with
novel objects, in novel scenes, and from novel viewpoints.
Our approach anchors generation to rendered robot motion while conditioning
on explicit scene and object priors, effectively decoupling trajectory
execution from environment synthesis.
This formulation has the potential to unlock two powerful data scaling capabilities:
(1)~\emph{retrieval and rebirth}, which repurposes existing
trajectories into entirely new contexts without new motion data; and
(2)~\emph{prop-free teleoperation}\footnote{The term “prop-free” is borrowed from film and television, where actors perform without physical props and the objects are later added via visual effects. In our context, it refers to teleoperating without physical objects.}, where operators manipulate empty air
and the model hallucinates the target objects and scene afterwards,
eliminating reset time.
We demonstrate with real-world experiments that our generated data
consistently improves downstream policy performance and significantly
reduces real-world data requirements across diverse manipulation tasks. Our project page is available at \texttt{https://junjieye.com/RoboDream/}.

\end{abstract}

\section{Introduction}

Data is the fuel of modern robot learning.
While scaling laws have driven remarkable progress in computer vision and
natural language processing, robotic manipulation remains constrained by the
scarcity of high-quality interaction data~\cite{rt2,octo,lin2024data, khazatsky2024droid}.
Imitation learning, the dominant paradigm for acquiring manipulation skills,
typically relies on teleoperated demonstrations~\cite{intelligence2025pi05visionlanguageactionmodelopenworld, lbmtri2025, chi2023diffusion,zhao2023act}.
However, collecting such data at the scale required for general-purpose robots
is prohibitively expensive, labor-intensive, and difficult to diversify across
environments and objects.
Consequently, developing scalable ``data engines" that can synthesize useful
training data from limited real-world experience has become a central pursuit
in robotics.

Generative models, particularly video diffusion models trained on internet-scale
data, offer a promising avenue for this purpose~\cite{brooks2024sora,wan2025}.
By internalizing priors about physics and object interactions, these models can
simulate realistic robot behaviors.
However, transforming these priors into effective training data is non-trivial.
Early attempts focused on visual augmentation~\cite{rosie,roboengine}, which
diversifies textures and backgrounds but fails to generate novel physical
interactions.
More recent approaches generate full video demonstrations from text and
extract actions via inverse dynamics~\cite{dreamgen, Ye2026DreamZero}.
While scalable, these methods often suffer from embodiment inconsistency,
where the generated robot morphology or dynamics deviate from reality, leading
to policy failure on physical hardware.

To address embodiment consistency, prior work such as AnchorDream~\cite{anchordream} proposed
grounding the generation process in robot motion.
By conditioning the video model on a rendered video of the robot's kinematic
trajectory, AnchorDream ensures that the generated demonstrations are physically
feasible and kinematically accurate.
While effective for expanding existing datasets, AnchorDream's utility is
bounded by its reliance on fine-tuning.
To synthesize valid interactions, the model typically requires fine-tuning on a
set of in-domain demonstrations for each specific task or environment.
This creates a practical dependency: generating data for a new scenario first
requires collecting real data from that scenario to adapt the model.
Additionally, because the model learns the environment distribution implicitly
from this data, it offers limited explicit control over the generated scene or
objects, making it challenging to systematically vary environmental factors for
robustness.

We argue that overcoming these constraints requires understanding the
\emph{fundamental compositional nature of manipulation}, that actions, objects,
and scenes are distinct, recombinable elements.
A robust data engine should be able to ``inpaint" arbitrary objects and scenes
around a valid robot motion based on semantic and visual cues, rather than
memorizing specific dataset distributions.
Achieving this disentanglement necessitates training on data with sufficient
diversity~\cite{khazatsky2024droid} to cover a vast range of object interactions and environmental
conditions.

As shown in Fig.~\ref{fig:teaser}, in this work we present \textbf{\methodname}, a generalizable embodiment-centric
world model that realizes this vision.
Our key insight is to decouple the generation of the robot's motion from the
generation of its context.
\methodname conditions the diffusion process on three explicit inputs:
(1) a rendered robot-only trajectory that anchors the embodiment;
(2) an \emph{object prior} that specifies the visual appearance of the target object;
and (3) a \emph{scene prior} that defines the background environment.
This formulation allows the model to ``paint" arbitrary objects and scenes around
a valid robot motion.
We scale up the training of \methodname with extensive multi-environment
data~\cite{khazatsky2024droid}, which enables the model to separate the robot's
kinematics from the visual context.
This scale-up unlocks the ability to synthesize valid interactions in zero-shot
scenarios unseen during training.
Consequently, unlike prior methods, it does not require task-specific fine-tuning
to generalize to new visual contexts; it can synthesize photorealistic
demonstrations with novel objects and scenes in a zero-shot manner.

This disentangled conditioning unlocks powerful new data generation capabilities.
Existing trajectories can be retrieved and ``reborn" in entirely new
contexts—new viewpoints, new lighting, and new background clutter—without
collecting a single new demonstration, a mode we term \emph{retrieval and rebirth}.
Furthermore, we enable \emph{prop-free teleoperation}, a streamlined data
collection workflow where operators manipulate empty air, and the model
hallucinates the target objects and interactions afterwards.
This eliminates the need for time-consuming physical resets between trials,
accelerating the wall-clock speed of data collection.

Our contributions are summarized as follows:
\begin{itemize}
    \item We propose \methodname, a controllable video world model that integrates
    explicit scene and object priors to achieve scalable and generalizable
    robot data synthesis.
    \item We introduce two novel data generation paradigms with \methodname, including
    retrieval and rebirth and prop-free teleoperation, which leverage
    the model's \textbf{zero-shot} generalization capabilities to multiply data diversity.
    \item We demonstrate through a series of real-world experiments that
    \methodname can generate high-quality data with novel objects and scenes
    from novel viewpoints in a \textbf{zero-shot} manner, consistently empowering better
    downstream policy performance compared to baselines.
\end{itemize}

\begin{figure*}[!t]
  \centering
    \includegraphics[width=0.97\linewidth]{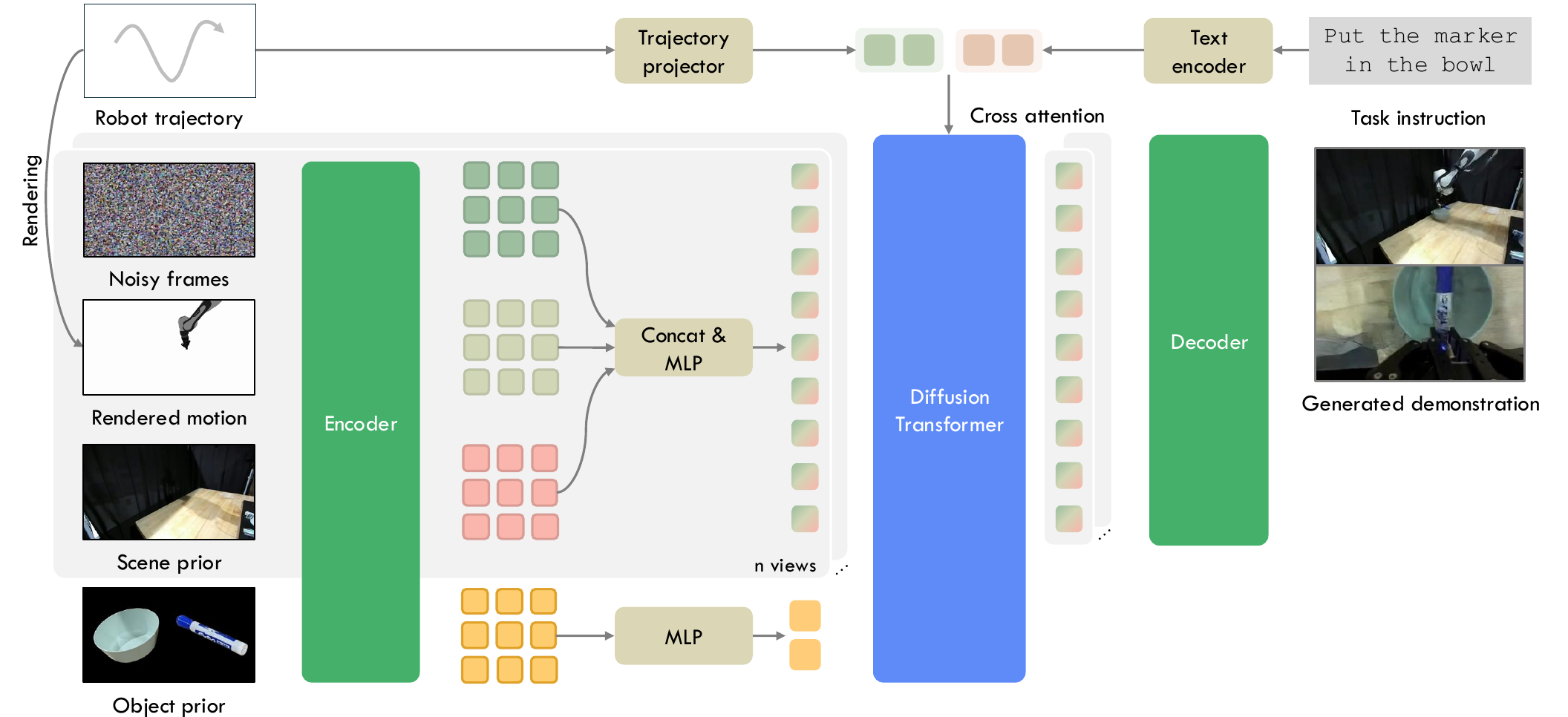}
  \caption{%
    The \methodname architecture.
    The model generates a video $o_{1:T}$ conditioned on multiple inputs: a rendered robot-only video $v_{\text{rob}}$ (kinematic anchor), a scene prior $I_s$ (background), and an object prior $I_o$ (visual appearance).
    We concatenate the latent video features with the encoded robot motion and scene prior, while injecting the object prior tokens via extended self-attention.
    Additionally, the task instruction and global robot trajectory are injected via cross-attention to ensure semantic and kinematic consistency.
  }
  \label{fig:pipeline}
 \vspace{-18pt}
  \end{figure*}

\section{Related Work}

\subsection{Generative World Models for Robot Learning}

The rapid advancement of generative video models has shifted focus from traditional video prediction~\cite{nvidia2025cosmosworldfoundationmodel} to large-scale video generation~\cite{wan2025,brooks2024sora}.
In robotics, these models serve as foundation world models capable of simulating diverse physical interactions.
DreamGen~\cite{dreamgen} generates video demonstrations from text and extracts policies via inverse dynamics.
Similarly, DreamDojo~\cite{gao2026dreamdojo} and DreamZero~\cite{Ye2026DreamZero} leverage large-scale human videos to learn generalist world models for zero-shot policy learning.
Cosmos-Predict2~\cite{nvidia_cosmos_predict2} provides a unified platform for physical AI simulation.
However, most of these approaches generate the robot and environment jointly, leading to potential embodiment hallucinations where the robot's morphology or kinematics may be inconsistent.
AnchorDream~\cite{anchordream} addresses this by anchoring generation to robot motion, but it lacks the generalization capabilities to zero-shot environments and objects that we achieve in \methodname.


\subsection{Scalable Data Synthesis \& Augmentation}

\textbf{Visual Augmentation.}
A significant body of work focuses on augmenting the visual diversity of existing demonstrations~\cite{chen2024roviaug, ji2025oxeauge} while keeping the underlying robot trajectory fixed.
ROSIE~\cite{rosie} employs text-guided inpainting to introduce diverse backgrounds and distractors into training data.
RoboEngine~\cite{roboengine} advances this by combining semantic robot segmentation with generative background synthesis to ensure better consistency.
Recent methods like RoboVIP~\cite{Wang2026RoboVIP} and RoboTransfer~\cite{liu2025robotransfer} utilize video diffusion models to enforce geometric consistency across multi-view augmentations.
Others, such as ReBot~\cite{fang2025rebot} and World Simulation~\cite{Ali2025WorldSimulation}, leverage video foundation models for real-to-sim-to-real transfer. RoboSplat~\cite{Yang2025RoboSplat} uses Gaussian Splatting for robust novel view synthesis, while AnyView~\cite{anyview2026} leverages implicit representations to generate dynamic scenes from arbitrary viewpoints.
While effective at improving visual robustness, these ``in-place" augmentation methods are fundamentally constrained by original demonstrations; they cannot generate new physical configurations or interaction layouts from scratch, limiting their ability to expand the behavioral distribution.
\methodname overcomes this limitation by decoupling motion from visual context, enabling the synthesis of entirely new physical configurations beyond original demonstrations.

\textbf{Generating New Physical Configurations.}
To go beyond visual diversity, several approaches synthesize entirely new trajectories and scenes.
MimicGen~\cite{mimicgen} procedurally generates new task demonstrations by composing sub-trajectories and rendering them in a simulator with varied object poses.
Real2Render2Real~\cite{r2r2r} reconstructs 3D assets from human videos to render diverse robot trajectories.
DemoGen~\cite{demogen} adopts a similar strategy using point cloud editing to recombine object-centric motions.
While these methods successfully expand the distribution of physical configurations, they rely heavily on explicit simulation assets, 3D object meshes, or digital twins, which are costly to acquire and difficult to scale to in-the-wild objects.
\methodname offers a distinct advantage by enabling the synthesis of new physical configurations—such as our prop-free teleoperation mode—directly in the visual domain.
By conditioning on explicit object and scene priors, we achieve the structural control of simulation-based methods with the visual realism and scalability of generative models, without requiring explicit 3D assets.

\section{Methodology}

\methodname is a generalizable world model designed to synthesize photorealistic
robot demonstrations by anchoring generation to robot motion while offering
explicit control over the environment and objects.
Our design is driven by the insight that actions, objects, and scenes are
distinct, recombinable elements in manipulation tasks.
By decoupling these components, \methodname can ``paint" arbitrary objects
and scenes around a valid kinematic trajectory without requiring task-specific
fine-tuning.
As illustrated in Fig.~\ref{fig:pipeline}, the model takes three key inputs:
(1) a rendered robot-only motion video that ensures kinematic feasibility;
(2) a scene prior image representing the background; and
(3) an object prior image containing task-relevant objects.
Trained on large-scale, diverse robot data, \methodname learns to
synthesize consistent interactions between the robot, objects, and scene,
enabling scalable data generation for zero-shot environments.

\subsection{Problem Formulation}

Let $\tau = \{(s_t, a_t)\}_{t=1}^T$ denote a robot trajectory, where
$s_t$ is the robot state and $a_t$ the action at step $t$, and let
$o_{1:T} = \{o_t\}_{t=1}^T$ be the corresponding visual observations.
Given a task description $\ell$, our goal is to synthesize a large dataset
$\mathcal{D}' = \{(\tau'_j, o_{1:T}'^j)\}$ with:
(i) diverse robot trajectories $\tau'_j$ consistent with task $\ell$,
(ii) realistic observations $o_{1:T}'$ consistent with $\tau'_j$, and
(iii) controllable environments and objects specified by external priors.

\methodname achieves this by conditioning video generation on three inputs:
a rendered robot motion video $v_{\text{rob}} \in \mathbb{R}^{T \times H \times W \times 3}$,
a scene prior image $I_s \in \mathbb{R}^{H \times W \times 3}$, and an
object prior image $I_o \in \mathbb{R}^{H \times W \times 3}$.
Note that $v_{\text{rob}}$ implicitly captures camera information by depicting the robot's motion in the pixel domain.
Consequently, by rendering the same trajectory from a different viewpoint and coupling it with a corresponding scene prior $I_s$, we can generate demonstrations at novel views.
Furthermore, both $I_s$ and $I_o$ can be replaced with arbitrary images
at inference time, enabling generalization to unseen environments and
objects without retraining.

\subsection{Model Architecture}

The core of \methodname is a multi-modal video diffusion transformer tailored
for embodiment-aware synthesis.
We approximate the conditional distribution:
\begin{equation}
    p_\theta(o_{1:T} \mid v_{\text{rob}}, I_s, I_o, \ell, \tau)
\end{equation}
where $\ell$ denotes the language instruction and $\tau$ the global trajectory context.
We introduce several architectural innovations to handle these distinct conditioning signals effectively.

\textbf{Multi-Modal Channel Extension.}
We extend the input channel of the diffusion backbone to accept a multi-modal
tensor.
Specifically, we concatenate the noisy video latent frames $z_t$ with the VAE-encoded
rendered robot motion video $\mathcal{E}(v_{\text{rob}})$ and the encoded scene prior $\mathcal{E}(I_s)$.
Since the scene background is largely static throughout the sequence, we
expand the single scene prior image $I_s$ along the temporal dimension to
match the video length $T$, treating it as a static video $I_s^T$.
The joint input to the transformer is formulated as:
\begin{equation}
    x_{\text{in}} = \text{Concat}(z_t, \mathcal{E}(v_{\text{rob}}), \mathcal{E}(I_s^T))
\end{equation}
This provides the model with a strong, pixel-aligned reference for both the
robot's intended motion and the environmental layout at every timestep.

\textbf{Multi-View Tokenization.}
To support multi-view generation without the spatial ambiguity of simple
concatenation, we adopt a token-based approach.
Instead of spatially concatenating frames from different cameras (\textit{e.g.}, side
view and wrist view) into a single wide image, we treat each view as a
separate video entry.
We independently tokenize the combined input $x_{\text{in}}$
for each view and then stack all tokens together before passing them to the
transformer.
This allows the model to process multiple viewpoints efficiently while
maintaining their distinct geometric perspectives.

\textbf{Object Prior via Self-Attention.}
The object prior $I_o$ consists of task-relevant objects randomly placed on a
single blank canvas.
This image is encoded using the same VAE as the video frames, producing a
set of latent object tokens $z_{\text{obj}} = \mathcal{E}(I_o)$.
These tokens are injected directly into the self-attention mechanism of the
transformer.
Formally, if $z_{\text{vid}}$ represents the sequence of video tokens derived from $x_{\text{in}}$, the
self-attention layers operate over the concatenated sequence:
\begin{equation}
    \text{Attention}(Q_{\text{vid}}, [K_{\text{vid}}; K_{\text{obj}}], [V_{\text{vid}}; V_{\text{obj}}])
\end{equation}
where $K_{\text{obj}}$ and $V_{\text{obj}}$ are keys and values projected from $z_{\text{obj}}$.
This allows the model to attend to the visual details of the objects at any
point in the video generation process, facilitating accurate object inpainting
regardless of where the objects appear in the scene.

\textbf{Cross-Attention Conditioning.}
In addition to visual priors, we condition the model on the task description
and global trajectory structure.
The text instruction $\ell$ is encoded via a T5 text encoder~\cite{t5encoder}.
Following AnchorDream~\cite{anchordream}, we also encode the global trajectory
states $\tau$ using an MLP.
These embeddings are injected via cross-attention layers, guiding the
high-level semantic and kinematic consistency of the generation.

\begin{figure}[!t]
  \centering
    \includegraphics[width=0.98\linewidth]{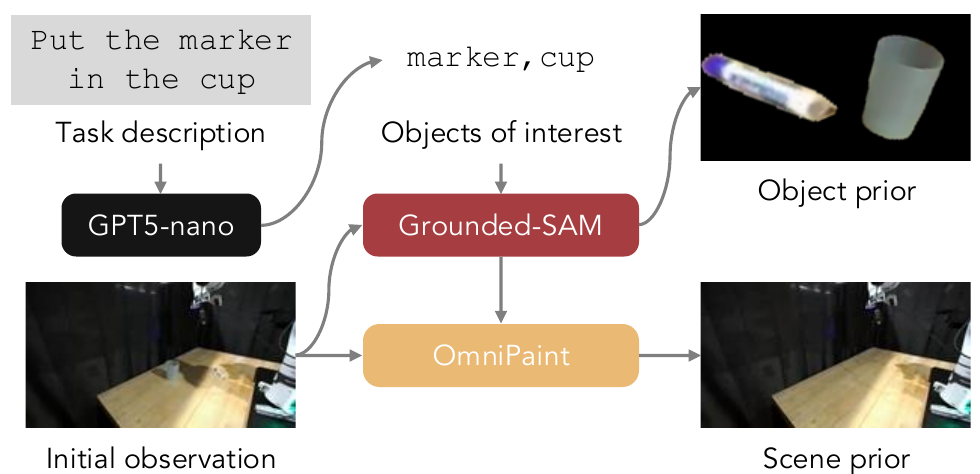}
  \caption{Data processing stages for extracting scene priors (background after object removal via OmniPaint) and object priors (cropped objects composited into a single image via Grounded-SAM).}
  \label{fig:data-processing}
 \vspace{-15pt}
  \end{figure}
  
\subsection{Prior Extraction}

As shown in Fig.~\ref{fig:data-processing}, to train \methodname, we construct training pairs $(v, I_s, I_o)$ from
existing large-scale robot datasets~\cite{khazatsky2024droid}.
This automated pipeline extracts priors without manual annotation.

\textbf{Object Identification.}
Given the first frame $o_1$ and task instruction $\ell$, we use a
vision-language model (GPT-5-nano~\cite{openai2025gpt5}) to identify the objects explicitly
relevant to the task.
The model returns a list of object names (\textit{e.g.}, ``red cup," ``blue sponge"),
filtering out background elements like tables or walls.

\textbf{Object Prior Construction.}
We apply Grounded-SAM~\cite{groundedsam} to segment the identified objects
in $o_1$.
To create the object prior image $I_o$, we crop the segmented objects,
randomly rotate and scale them, and place them on a clean canvas.
This random placement prevents the model from overfitting to the objects'
original positions, encouraging it to learn their visual appearance
independent of their location.

\textbf{Scene Prior Construction.}
We generate the scene prior $I_s$ by removing the segmented task-relevant
objects from the first frame $o_1$.
We utilize OmniPaint~\cite{omnipaint}, a diffusion-based inpainting model, to
fill in the holes left by object removal, resulting in a clean background image
of the workspace.

\subsection{Deployment Modes}

\textbf{Retrieval and Rebirth.}
Given a new task, we retrieve semantically similar trajectories from an
existing dataset.
To do this, we use a T5 encoder~\cite{t5encoder} to embed the query task instruction and compute its cosine similarity with the instruction embeddings of all available trajectories, selecting the top matches.
These trajectories are replayed in a simulator (Isaac Lab) to render
robot-only motion videos $v_{\text{rob}}$ from novel camera viewpoints.
We then combine $v_{\text{rob}}$ with new scene and object priors to synthesize
demonstrations reborn in novel contexts.

\textbf{Prop-Free Teleoperation.}
We introduce a novel data collection workflow where the teleoperator
controls the robot to perform task motions with imaginary objects (pantomime).
Crucially, this collection can occur either in the \textbf{real world} (using an empty workspace)
or directly in a \textbf{simulator}, as \methodname requires only the valid kinematic trajectory,
not physical contact data.
The recorded trajectory is rendered to produce the robot motion video $v_{\text{rob}}$.
\methodname then combines this motion with arbitrary target object ($I_o$)
and scene ($I_s$) priors to synthesize a realistic video of the robot
interacting with the object.
This mode decouples data collection from physical constraints and specific environments, allowing for
rapid, continuous teleoperation without the need for time-consuming scene resets
or precise object manipulation, as the model handles the fine-grained
interaction details.

\section{Experiments}

We evaluate \methodname on a suite of four challenging everyday manipulation
tasks in the real world: (1)~\emph{Put Marker into Bowl}, (2)~\emph{Remove Marker from Bowl},
(3)~\emph{Put Cube into Cup}, and (4)~\emph{Wipe Table with Towel}.
Our experiments are designed to investigate:
(1) whether generated data can improve policy performance;
(2) the efficiency and effectiveness of prop-free teleoperation;
(3) the scaling properties of our data engine; and
(4) the model's zero-shot generalization capabilities.

\subsection{Experimental Setup}

\begin{figure}[!t]
  \centering
    \includegraphics[width=0.98\linewidth]{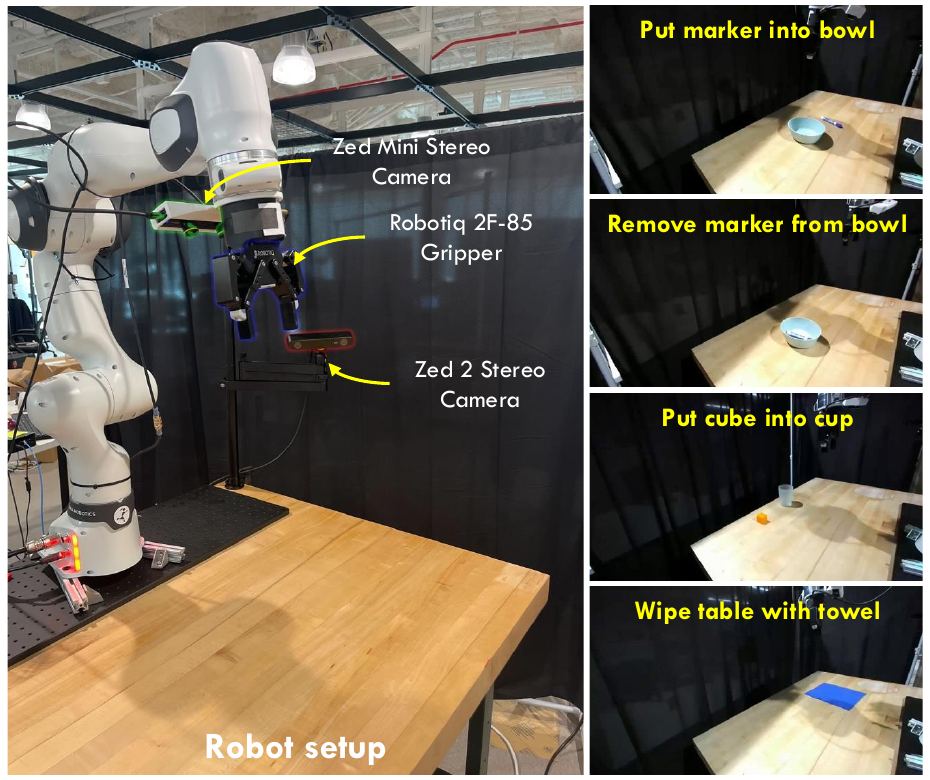}
  \caption{Real-world evaluation setup using the DROID robot platform.
  We evaluate on four tasks: \emph{Put Marker into Bowl}, \emph{Remove Marker from Bowl},
  \emph{Put Cube into Cup}, and \emph{Wipe Table with Towel}.}
  \label{fig:real_robot_setup}
 \vspace{-5pt}
  \end{figure}

\textbf{Task Setup.}
We conduct experiments on a Franka Panda robot setup consistent with the
DROID~\cite{khazatsky2024droid} platform, as shown in Fig.~\ref{fig:real_robot_setup}.
Each task involves precise object manipulation.
We evaluate success based on task completion; for pick-and-place tasks, we
consider partial success (\textit{e.g.}, successful grasp but failed placement) as half credit.
All real-world evaluations are conducted over 20 rollouts per policy.

\textbf{Training Setup.}
\methodname is fine-tuned from the Cosmos-Predict2 2B~\cite{nvidia_cosmos_predict2} foundation model.
We train on approximately 40k episodes from the DROID dataset where camera
calibration is available.
Training is performed on 2 nodes of 8 NVIDIA A100 GPUs for one week.
The model processes observations from two cameras: a third-person static camera
and a wrist-mounted camera.
For robot-only video rendering, we utilize Isaac Lab to generate consistent
kinematic motion traces.
Downstream policies are trained using Diffusion Policy~\cite{chi2023diffusion}
to isolate and examine the effect of \methodname-generated data on policy learning.

\subsection{Retrieval and Rebirth for Policy Learning}

We first investigate whether retrieving and rebirthing demonstrations from
DROID can improve policy performance in our target domain.
We compare five data regimes to isolate the contribution of generated data.
\textbf{Real-50} uses 50 in-domain real demonstrations (collected across the 3 pick-and-place tasks).
\textbf{Gen-100} uses 100 demonstrations retrieved from DROID and reborn in our setting.
\textbf{Gen-Mix} combines real and generated data with 50\% sampling probability.
We also evaluate baselines using raw retrieved DROID data:
\textbf{Orig-100} uses the original retrieved episodes (without rebirth), and
\textbf{Orig-Mix} mixes them with Real-50.
We exclude prior motion-conditioned approaches~\cite{anchordream} from this comparison as they lack the explicit conditioning mechanisms necessary to generate demonstrations in zero-shot environments without in-domain fine-tuning.

\begin{table}[!t]
\caption{Policy success rate (\%) across data regimes. \textbf{Gen-Mix} achieves the best performance, significantly outperforming baselines.}
\label{tab:main}
\begin{center}
\resizebox{\columnwidth}{!}{%
\begin{tabular}{lccccc}
\toprule
Task & Real-50 & Orig-100 & Orig-Mix & Gen-100 & Gen-Mix \\
\midrule
Put Cube into Cup       & 35 & 0 & 55 & 20 & \textbf{65} \\
Put Marker into Bowl    & 30 & 0 & 35 & 15 & \textbf{55} \\
Remove Marker from Bowl & 20 & 0 & 20 & 5  & \textbf{35} \\
Wipe Table with Towel   & 60 & 0 & 70 & 20 & \textbf{95} \\
\midrule
Average                 & 36.3 & 0 & 45.0 & 15.0 & \textbf{62.5} \\
\bottomrule
\end{tabular}%
}
\end{center}
 \vspace{-15pt}
\end{table}

  \begin{figure*}[!t]
  \centering
    \includegraphics[width=0.98\linewidth]{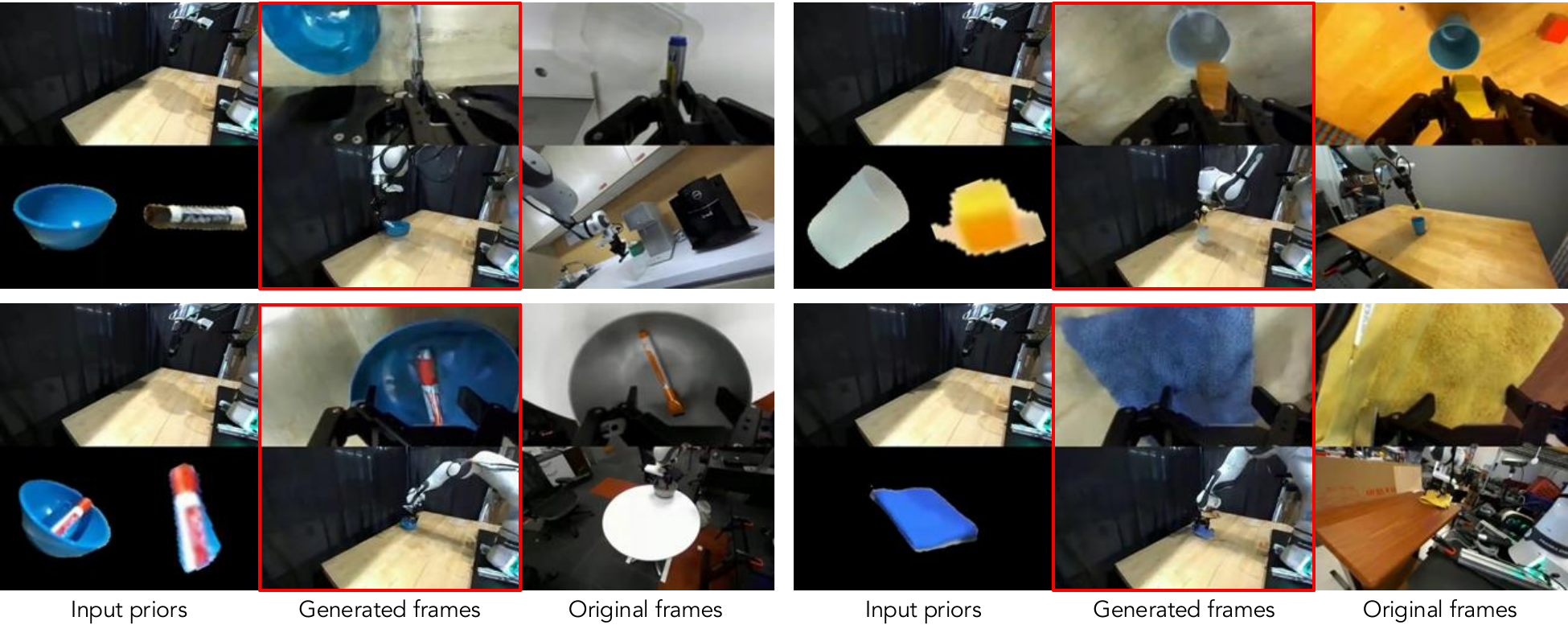}
  \caption{%
      Zero-shot demonstration rebirth with \methodname.
      We retrieve existing trajectories from the DROID dataset (Right) and
      rebirth them in our target setting.
      Given new object priors and scene priors (Left), the model synthesizes
      photorealistic demonstrations (Middle) where robot interacts with
      specified objects in the specified scene, adhering to the retrieved motion.
  }
  \label{fig:reborn-existing}
 \vspace{-10pt}
  \end{figure*}

As shown in Tab.~\ref{tab:main}, \textbf{Gen-Mix} consistently outperforms
all other methods, achieving an average success rate of 62.5\% compared to 36.3\%
for Real-50 and 45.0\% for Orig-Mix.
The failure of \textbf{Orig-100} (0\% success) highlights the significant
domain shift between DROID's diverse environments and our specific target setup.
Specifically, the viewpoints, scene layouts, and object instances in the retrieved DROID data are distinct from our target environment, leading to a large covariate shift that the policy cannot bridge without adaptation.
In contrast, \methodname successfully bridges this gap.
Fig.~\ref{fig:reborn-existing} visualizes this ``rebirth" process: given
target object and scene priors (Left), the model transforms retrieved DROID
trajectories (Right) into photorealistic demonstrations in our setup (Middle).
Crucially, all objects and scenes used in these experiments are \textbf{unseen} during
training, demonstrating the model's zero-shot generalization capability.

\subsection{Prop-Free vs.\ Real Data Collection}

Next, we evaluate the effectiveness of our prop-free teleoperation mode,
which aims to reduce collection cost without compromising policy performance.
We compare prop-free teleoperation (motion collection + generation)
against standard real teleoperation in terms of efficiency and policy
performance. We also evaluate a \textbf{Real w/ Gen Obs} baseline, where we take real
trajectories and replace the observations with \methodname-generated ones, to
isolate the visual generation quality from trajectory quality.

\begin{table}[t]
  \caption{Policy success rate (\%) comparing prop-free teleoperation against real data collection baselines.}
  \label{tab:propfree}
  \begin{center}
  \small
  \resizebox{\columnwidth}{!}{%
  \begin{tabular}{lccc}
  \toprule
  Task & Real-50 & Real w/ Gen Obs & Prop-Free \\
  \midrule
  Put Cube into Cup       & 35 & 25 & 30 \\
  Put Marker into Bowl    & 30 & 20 & 20 \\
  Remove Marker from Bowl & 20 & 15 & 20 \\
  Wipe Table with Towel   & 60 & 60 & 60 \\
  \midrule
  Average                 & 36.3 & 30.0 & 32.5 \\
  \bottomrule
  \end{tabular}%
  }
  \end{center}
 \vspace{-10pt}
  \end{table}

\textbf{Efficiency.}
Collecting 50 real-world teleoperation episodes required approximately \textbf{2 hours}
due to the need for manual scene resets and object repositioning between trials.
In contrast, collecting 50 prop-free trajectories took only \textbf{55 minutes},
as the operator could continuously perform motions without interacting with
physical objects.
Notably, since pick-and-place tasks share similar motion patterns, we collected
a single pool of 50 trajectories in this prop-free setting and used \methodname
to paint corresponding objects for all three pick-and-place tasks.
While conservative (a single pattern may not perfectly generalize), this
shared-trajectory strategy further amplified the efficiency gains.

\textbf{Performance.}
Tab.~\ref{tab:propfree} compares the policy success rates.
\textbf{Real w/ Gen Obs} achieves performance very close to \textbf{Real-50} (average 30.0\% vs 36.3\%),
demonstrating the high fidelity of our visual generation.
Crucially, \textbf{Prop-Free} achieves competitive performance (average 32.5\%),
matching or slightly underperforming the real baseline while being $\sim 2.2\times$ faster to collect.
This verifies that \methodname enables scalable data collection without the
logistical bottleneck of physical resets.

\begin{table}[t]
  \caption{Policy success rate (\%) as generated data scales (mixed with Real-50). Performance improves with more data and saturates around Mix-200.}
  \label{tab:scaling}
  \begin{center}
  \small
  \resizebox{\columnwidth}{!}{%
  \begin{tabular}{lccccc}
  \toprule
  Task & Real-50 & Mix-100 & Mix-200 & Mix-300 & Mix-400 \\
  \midrule
  Put Cube into Cup       & 35 & 65 & 75 & 80 & 75 \\
  Put Marker into Bowl    & 30 & 55 & 70 & 70 & 70 \\
  Remove Marker from Bowl & 20 & 35 & 45 & 50 & 50 \\
  Wipe Table with Towel   & 60 & 95 & 100 & 95 & 100 \\
  \midrule
  Average                 & 36.3 & 62.5 & 72.5 & 73.75 & 73.75 \\
  \bottomrule
  \end{tabular}%
  }
  \end{center}
 \vspace{-10pt}
  \end{table}

\begin{figure*}[!t]
  \centering
    \includegraphics[width=0.99\linewidth]{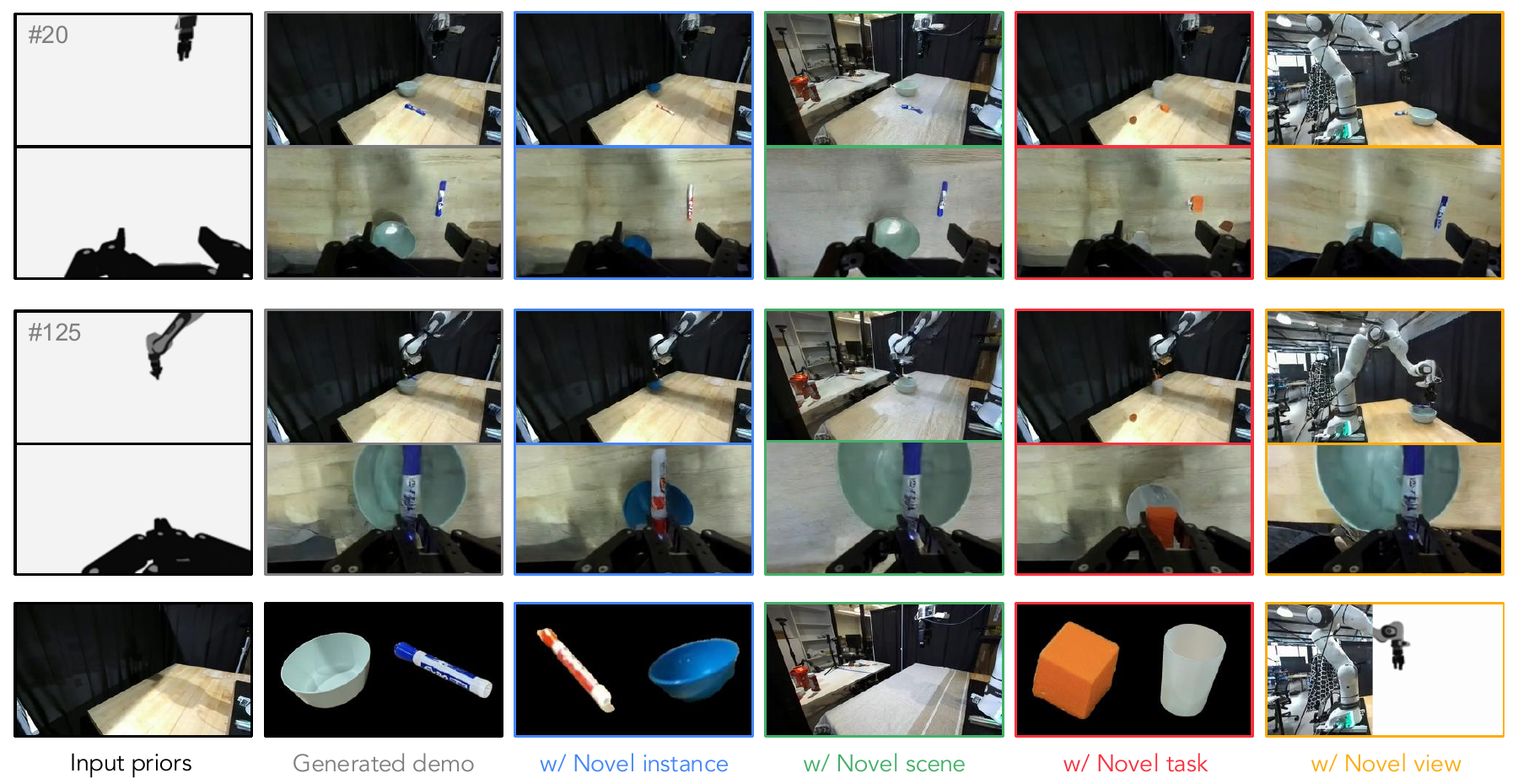}
  \caption{%
      Compositional generation with \methodname.
      We demonstrate our claim that actions, objects, and scenes are distinct,
      recombinable elements.
      Starting from a \textbf{\textit{single}} base trajectory, we can interpret it in multiple ways
      by changing the inputs:
      (1)~\textbf{Novel Instances}: Manipulating new instances (\textit{e.g.}, different markers) by changing the object prior.
      (2)~\textbf{Novel Scenes}: Placing the robot in a new environment by changing the scene prior.
      (3)~\textbf{Novel Tasks}: Changing the task context (\textit{e.g.}, marker-to-bowl vs cube-to-cup) by swapping object priors.
      (4)~\textbf{Novel Viewpoints}: Generating demonstrations from a new camera angle by re-rendering the motion and capturing a corresponding scene prior.
      All generations shown are \textbf{zero-shot}, using objects and scenes unseen during training.
  }
  \label{fig:compositionality}
  \vspace{-5pt}
  \end{figure*}

\subsection{Scaling up with Generated Data}

We further study the scaling properties of our data engine.
We fix the real dataset at 50 demonstrations and progressively increase
the number of generated demonstrations mixed in.
The generated data consists of retrieved DROID trajectories reborn in our
target setting.
Tab.~\ref{tab:scaling} reports success rates as generated data scales
from 100 up to 400 additional episodes.

We observe that adding generated data consistently improves
performance over the Real-50 baseline.
The performance gains appear to saturate at approximately \textbf{Mix-200}.
This suggests that while \methodname provides a scalable source of data,
the diversity of retrieved trajectories (from DROID) or the domain gap
in generation eventually limits the marginal benefit of adding more samples.
Nevertheless, the ability to significantly improve policy performance without collecting
additional real data demonstrates the power of our data engine.

\subsection{Compositional Generation}

Finally, we investigate the model's zero-shot generalization capabilities
through compositional generation.
As visualized in Fig.~\ref{fig:compositionality}, \methodname effectively
decouples the components of manipulation, allowing us to recombine them
to generate diverse demonstrations in a zero-shot manner.

First, we demonstrate generalization to \textbf{novel instances} by modifying the object prior $I_o$.
For instance, the model can substitute a blue marker with a red marker while preserving the underlying physical interaction fidelity.
Second, we evaluate the capability to generalize to \textbf{novel scenes} by altering the scene prior $I_s$.
Leveraging the diversity of the DROID training set, \methodname seamlessly integrates the robot into unseen backgrounds without generating visual artifacts, confirming its robust disentanglement of foreground and background elements.
Third, we synthesize \textbf{novel tasks} (\textit{e.g.}, transitioning from picking a marker to picking a cube) using the same kinematic trajectory by supplying an alternative object prior, provided that the grasp affordances remain compatible.
This underscores the model's capacity for object-centric manipulation reasoning.
Finally, by re-rendering the robot-only motion video $v$ from an alternative camera pose and acquiring a corresponding scene prior $I_s$, we achieve synthesis from \textbf{novel viewpoints}.
This capability facilitates the training of robust multi-view policies derived from single-view data sources.

\section{Limitations}

\methodname assumes that the rendered robot-only video faithfully captures
the kinematic trajectory that the robot should follow in the target scene.
While generation quality depends on the coverage of the training distribution,
we believe this can be resolved by further scaling up training with extensive data,
including human videos.
By treating the human as an embodiment and grounding generation to human motion,
our framework can potentially leverage internet-scale human data to improve robustness.
Additionally, the model inherits the temporal length and resolution
limitations of the underlying video diffusion backbone.
We expect these limitations to diminish as base video generation models continue to improve.


\section{Conclusion}

We presented \methodname, a generalizable embodiment-centric world model
for robot data synthesis.
By conditioning video generation on explicit scene and object priors,
our method enables controllable demonstration generation across arbitrary
environments and objects without task-specific fine-tuning.
Two practical deployment modes, retrieval and rebirth and
prop-free teleoperation, address key bottlenecks in real-world data
collection.
Experiments show that \methodname consistently improves policy performance
and reduces the burden of real data collection.
We believe this work provides a practical path toward scalable imitation
learning without the need for massive teleoperation efforts or task-specific
simulation setups.


\section{Acknowledgment}
We are grateful to Jiawei Yang, Mingtong Zhang, and Weiduo Yuan for their helpful discussions, and to Abrar Anwar and Jesse Thomason for their support with the robot hardware.
The USC Physical Superintelligence Lab acknowledges generous support from Toyota Research Institute, Dolby, Google DeepMind, Capital One, Nvidia, Bosch, NSF, and Qualcomm. Junjie Ye is supported by a Capital One Fellowship. Yue Wang is also supported by a Powell Research Award.


\bibliographystyle{IEEEtran}
\bibliography{iros2026}

\end{document}